\documentclass[letterpaper,10pt,conference]{ieeeconf} 
\usepackage{cite}
\usepackage{amsmath,amssymb,amsfonts}
\usepackage{algorithmic}
\usepackage{graphicx}
\usepackage{textcomp}
\usepackage{xcolor}
\usepackage{float}
\usepackage{url}
\usepackage{booktabs}
\usepackage{caption}
\usepackage{stfloats}
\usepackage{makecell}

\IEEEoverridecommandlockouts                              

\usepackage{url}
\usepackage{subfig}
\usepackage{graphicx}
\usepackage{varioref}
\graphicspath{{figures/}}
\labelformat{equation}{\textup{(#1)}}
\usepackage[pageanchor=true,
    plainpages=false,
    pdfpagelabels,
    bookmarks,
    bookmarksnumbered,
    pdftex,	
    pdfstartview=FitH,
    linkcolor=black,
    citecolor=black,
    urlcolor=black,
    filecolor=black]{hyperref}
\usepackage{amssymb}
\usepackage{multirow}
\usepackage{todonotes}
\usepackage{flushend}
\usepackage{tikz}

\usepackage{array}
\newcolumntype{L}[1]{>{\raggedright\let\newline\\\arraybackslash\hspace{0pt}}m{#1}}
\newcolumntype{C}[1]{>{\centering\let\newline\\\arraybackslash\hspace{0pt}}m{#1}}
\newcolumntype{R}[1]{>{\raggedleft\let\newline\\\arraybackslash\hspace{0pt}}m{#1}}

\def\usenatbib{1}
\ifx\usenatbib\undefined
    \usepackage{cite}
\else
    \makeatletter
    \let\NAT@parse\undefined
    \makeatother
    \usepackage[numbers,sort]{natbib}

    \setcitestyle{citesep={], [}}
    \makeatletter
    \def\NAT@def@citea{\def\@citea{\NAT@separator}}%
    \makeatother
\fi

\graphicspath{ {./images/} }

\let\orgautoref\autoref
\providecommand{\Autoref}
        {\def\equationautorefname{Equation}%
         \def\figureautorefname{Figure}%
         \def\subfigureautorefname{Figure}%
         \def\Itemautorefname{Item}%
         \def\tableautorefname{Table}%
         \def\exerciseautorefname{Exercise}%
         \def\starexerciseautorefname{Exercise}%
         \def\sectionautorefname{Section}%
         \def\subsectionautorefname{Section}%
         \def\subsubsectionautorefname{Section}%
         \def\chapterautorefname{Section}%
         \def\partautorefname{Part}%
         \orgautoref}

\renewcommand{\autoref}
        {\def\equationautorefname{Equation}%
         \def\figureautorefname{Fig.}%
         \def\subfigureautorefname{Fig.}%
         \def\Itemautorefname{item}%
         \def\tableautorefname{Table}%
         \def\exerciseautorefname{Exercise}%
         \def\starexerciseautorefname{Exercise}%
         \def\sectionautorefname{Section}%
         \def\subsectionautorefname{Section}%
         \def\subsubsectionautorefname{Section}%
         \def\chapterautorefname{Section}%
         \def\partautorefname{Part}%
         \orgautoref}

\usepackage{hyperref}
\hypersetup{
colorlinks=true,
urlcolor=red,
}
\title{\LARGE \bf 
Enabling 24-hour Agricultural Robotics: Unsupervised Day-to-Night Cross-Modal Image Translation for Nighttime Visual Navigation 
}

\author{Robel Mamo$^{1}$, Rajitha de Silva$^{2}$, Grzegorz Cielniak$^{2}$, and Taeyeong Choi$^{1}$
\thanks{\raggedright
        $^{1}$LaSER Lab, 
        Kennesaw State University, Marietta, GA 30060, USA. 
        {\tt\scriptsize  rmamo@students.kennesaw.edu, tchoi@kennesaw.edu}.
        \newline
        $^{2}$Lincoln Institute for Agri-Food Technology, 
        University of Lincoln, Lincoln LN2 2LG, UK. 
        \mbox{\tt\scriptsize  \{ODeSilva, GCielniak\}@lincoln.ac.uk}.} %
}

\begin{document}

\onecolumn
    \vspace*{\fill}
    This work has been accepted to the 2026 IEEE/RSJ International Conference on Intelligent Robots and Systems (IROS 2026). © 2026 IEEE. Personal use of this material is permitted. Permission from IEEE must be obtained for all other uses.
\vspace*{\fill}
\twocolumn
\clearpage

\maketitle
\thispagestyle{empty}
\pagestyle{empty}

\begin{abstract}

While visual navigation has been extensively studied in agricultural robotics,  most existing systems assume daytime conditions. 
In fact, deploying autonomous robots at night offers significant advantages, including $24$-hour crop and soil monitoring, fruit harvesting, and nocturnal pest detection. 
Modern vision-based systems, however, rely heavily on large-scale well-annotated image datasets, which remains challenging to obtain for nighttime operation scenarios.
To address this, we propose an unsupervised image translation framework that converts daytime plant-row RGB~images into near-infrared~(NIR) nighttime counterparts without requiring pixel-to-pixel supervision. 
This enables the direct \emph{reuse} of daytime semantic labels for training nighttime perception models. 
In particular, by incorporating a pre-trained Contrastive Language–Image Pretraining (CLIP) model, the proposed framework is designed to preserve semantic consistency during day-to-night translation.
Additionally, a visibility mask is introduced to account for the limited effective range of NIR illumination in nighttime scenes.
We conduct comparative evaluations with state-of-the-art image translation baselines and demonstrate higher image qualities, as supported by improved performance in downstream semantic segmentation for nighttime visual navigation. 
For evaluation, we utilize \emph{AgriNight}---a novel dataset comprising $428$~daytime and $549$~nighttime images collected using night-vision-equipped mobile robots in agricultural fields and manually annotated with pixel-wise semantic labels---and introduce it as the \emph{first} benchmark for nighttime agricultural visual navigation. 
We also perform real-time autonomous navigation experiments with a physical robot operating at night.
The data and code are available at: \href{https://github.com/mamorobel/AgriNight}{\textcolor{blue}{https://github.com/mamorobel/AgriNight}}.

\end{abstract}

\section{Introduction}
\label{sec:introduction}

Autonomous navigation is a fundamental capability for mobile robots operating in agricultural environments, enabling tasks such as crop monitoring, pest detection, precision spraying, and harvesting. 
Over the past decade, visual navigation has been extensively studied in the agricultural robotics community to allow robots to perceive their surroundings and traverse diverse crop fields safely and reliably~\cite{SGMHAC24, DCWG24}. 
Despite these advances, most existing systems have been developed and evaluated under daytime conditions. 
In contrast, \emph{nighttime} navigation has been widely investigated in general on-road and off-road environments~\cite{LLTLGJXY24, DPNRRPZHWO25}, while it remains relatively underexplored in agricultural settings.

\begin{figure}[t]
    \centering
    \includegraphics[width=.95\linewidth]{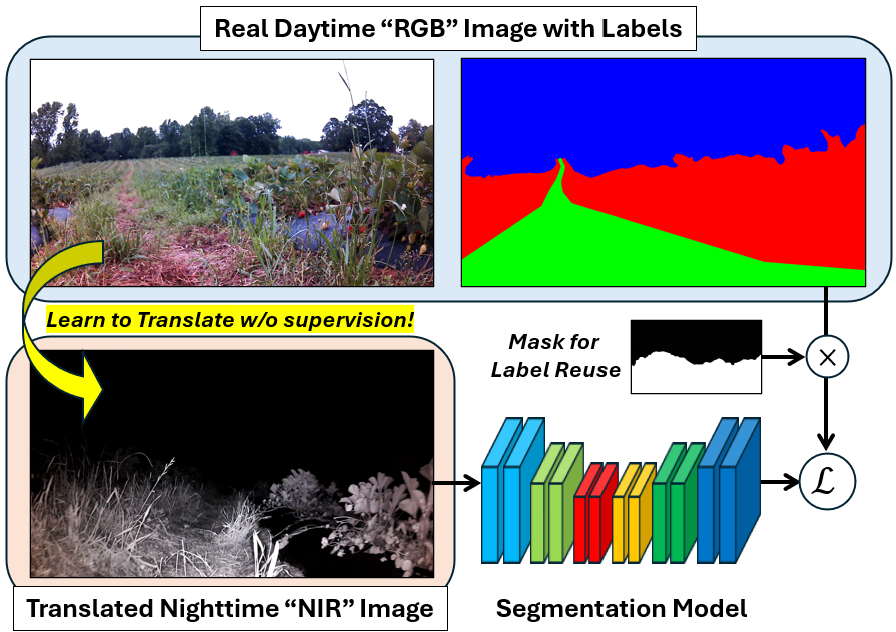}
    \caption{
    Illustration of the downstream workflow. Nighttime NIR images translated from real daytime RGB images are used to train the segmentation model. Daytime labels are reused with a mask to simulate limited nighttime visibility.} 
    \label{fig:concept}
\end{figure}

In fact, nighttime deployment of mobile robots can expand the operational bandwidth of agricultural automation~\cite{DPBGCCCDDF18}, enabling more continuous and efficient food production. 
For example, extended operational availability would allow robots to complete harvesting within a narrow time window ensuring optimal fruit quality, 
even if their harvesting speed is slower than that of human workers~\citep{KWBHV21}.
Furthermore, cooler nighttime temperatures can help preserve the firmness and post-harvest quality of soft fruits~\cite{LBYSGK15}. 
Nighttime navigation would also enable robots to continuously monitor fields, detect nocturnal pests, and perform timely interventions.  

A key challenge for developing nighttime agricultural navigation systems is, however, the limited availability of annotated nighttime data, as state-of-the-art learning-based approaches require large volumes of labeled training data. 
Existing datasets~\cite{SGMHAC24, DCWG24} primarily consist of daytime RGB images and thus do not adequately represent nighttime field conditions. 
Collecting nighttime data is particularly challenging in agricultural fields due to limited visual feedback for operators and heightened safety concerns, while reduced illumination further complicates annotation by requiring additional effort for precise object identification.

To address this limitation, we propose an unsupervised image translation framework that converts daytime plant-row RGB~images into realistic near-infrared~(NIR) nighttime counterparts without requiring pixel-to-pixel supervision.
As shown in~\autoref{fig:concept}, this enables the direct \emph{reuse} of daytime semantic labels for training nighttime perception models. 
The framework incorporates a pre-trained ``Contrastive Language-Image Pretraining''~(CLIP) model~\citep{RKHRGASAMC21} to preserve semantic and structural consistency during translation. 
Additionally, a visibility mask is introduced to account for the limited effective range of NIR illumination in nighttime scenes. 
For evaluation, we train semantic segmentation models only with the translated images and validate them on real nighttime images, aiming to integrate the resulting models into a navigation controller. 
We further demonstrate their effectiveness through physical robot experiments under nighttime conditions.

We utilize \emph{AgriNight}---a novel dataset consisting of $428$~daytime RGB and $549$~nighttime NIR images collected using our night-vision-equipped mobile robot from strawberry and carrot fields and manually annotated with pixel-wise semantic labels---and introduce it as the \emph{first} benchmark for nighttime agricultural visual navigation.  
In particular, we adopt $850$~nm NIR illumination for nighttime imaging rather than visible light, considering the following advantages.
First, it lies beyond plant photoreceptors' sensing range~\citep{GF15}, minimizing disruption to growth~\citep{MOYATS17}.
Second, unlike visible light, it attracts fewer nocturnal insects that could disrupt on-board perception.
Third, it yields high soil-vegetation contrast, which is critical for navigation, since chlorophyll-bearing foliage reflects more strongly than soil~\citep{T79}.

Our contributions are summarized as follows:
\begin{itemize}
\item A CLIP- and visibility mask-enhanced unsupervised image translation framework for cross-modal (RGB-to-NIR) day-to-night domain transfer in agricultural fields;
\item Comprehensive evaluation demonstrating superior performance over state-of-the-art image translation methods, including real-world robot experiments; and 
\item The AgriNight dataset for benchmarking nighttime agricultural field navigation.
\end{itemize}
To the best of our knowledge, this work is the \emph{first} to present a perception pipeline specifically designed for nighttime agricultural robot navigation, along with a day/night dataset.


\section{Related Works}
\label{sec:related_work}



\subsection{Unsupervised Cross-Modal Image Translation}

Image translation between RGB imagery and other sensing modalities (e.g., thermal infrared (TIR)) has been extensively studied to leverage abundant RGB annotations for training models in target modalities where labeled data are scarce~\citep{LJCK23, SSAS24}. 
In particular, unsupervised image translation approaches do \emph{not} require paired images captured from the source and target modalities of the same scene. 
For instance, \mbox{SSL-RGB2IR}~\citep{SSAS24}, similar to~\citep{LJCK23}, performs unsupervised learning to convert RGB~images into corresponding TIR~images.
These pseudo RGB-TIR pairs are then subsequently used to train a refined image translator in a supervised manner. 
This approach enables the supervised learning of downstream perception tasks, such as object detection and semantic segmentation, as demontrated in urban driving scenarios by using the translated TIR imagery.

While our downstream task also involves semantic segmentation, several key distinctions set our work apart: 1)~we \emph{jointly} address cross-modal (i.e.,~RGB to NIR) and ``cross-illumination'' (i.e.,~day-to-night) translation to enable nighttime agricultural operation; 2)~our focus is on relatively unstructured agricultural environments---characterized with non-uniform crop spacing and diverse plant morphologies---rather than structured urban scenes (e.g.,~vehicles, buildings); and 3)~we validate the framework through a physical mobile robot operating in crop fields at night.




\subsection{Unsupervised Domain Adaptation in Agriculture}

In agricultural applications, unsupervised domain adaptation~(UDA) is a crucial capability for adapting perception models to novel environmental conditions, such as varying soil types, crop growth stages, and lighting. 
Previous works~\citep{GLWPS20, MWGLBPS23} performed unsupervised domain transfer from one field to another to enable semantic segmentation of crops, weeds, and soil without requiring target-domain labels. 
In particular, the framework in~\citep{GLWPS20} is designed to enhance CycleGAN-based generative models by jointly training a semantic segmentation network to preserve semantic consistency during translation.
However, such joint optimization may lead to \emph{unstable} supervision, particularly in early training stages when segmentation predictions are unreliable.
To mitigate this issue, our framework instead leverages a fixed, pretrained CLIP model~\citep{RKHRGASAMC21} to compute a semantic preservation loss in a shared embedding space. 

In addition, diverse agricultural contexts have been considered for UDA, including harvesting~\citep{SSN19}, leaf localization~\citep{TNG25}, growth stage prediction~\citep{GL24}, and plant row detection~\citep{DJPMG22}. 
These studies, however, omitted physical robot demonstrations and primarily focused on a single data modality (i.e.,~RGB imagery) for daytime scenarios. 
In contrast, our work addresses cross-modal translation (RGB to NIR) and domain adaptation (Day to Night)  simultaneously, validated also through physical robot deployment in the field. 


\subsection{Visual Navigation in Low-light Condition}

While visual navigation under low-light conditions has been investigated extensively in urban streets (e.g.,~illumination enhancement~\citep{LLTLGJXY24} and day-to-night translation~\citep{PPNZY24, WCCCLSL22}), agricultural research remains underexplored. 
The M2P2 dataset, collected in forested environments~\citep{DPNRRPZHWO25}, represents a step toward off-road low-light navigation. 
However, its primary focus lies in utilizing multi-modal sensory inputs, such as stereo RGB, event, and thermal cameras, along with high-resolution LiDAR, to supply high-throughput geometric information to the robot controller. 
Our approach instead aims to enable robust navigation in dark agricultural fields using only a single NIR night-vision camera.


\subsection{Nighttime Operation in Agricultural Robotics}

Most existing nighttime perception approaches in agriculture have relied on RGB sensors paired with high-intensity light sources from bulky external rigs to ensure target objects remain visible within the RGB spectrum~\citep{LYHYZA24, WBB24} 
As an alternative, researchers in~\citep{TOT25} employed TIR imagery with depth estimation, mitigating the effects of varying lighting conditions. 
Still, all these studies were predominantly conducted in the context of nighttime fruit harvesting robots. 

In contrast, our work addresses autonomous navigation in agricultural fields~\citep{DCWG24, SGMHAC24} by leveraging NIR night-vision sensors. 
As an alternative to TIR cameras, NIR cameras offer a \emph{low-cost}, high-resolution solution, preserving fine texture and edges. 
Recently, \citet{CSNSN24} deployed NIR cameras mounted on a mobile robot platform in agricultural fields at night. 
However, their focus was on classifying terrain conditions while the operator manually controlled the robot.

\begin{figure*}[t]\centering
    \subfloat[]{\label{fig:cycle1}%
    \includegraphics[width=.39\linewidth]{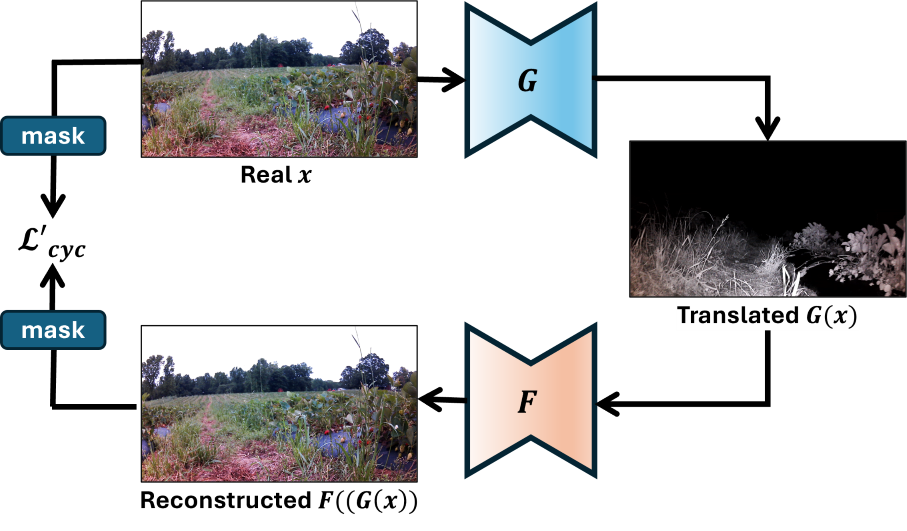}}
    \quad
    \quad
    \quad
    \subfloat[]{\label{fig:clip1}%
    \includegraphics[width=.39\linewidth]{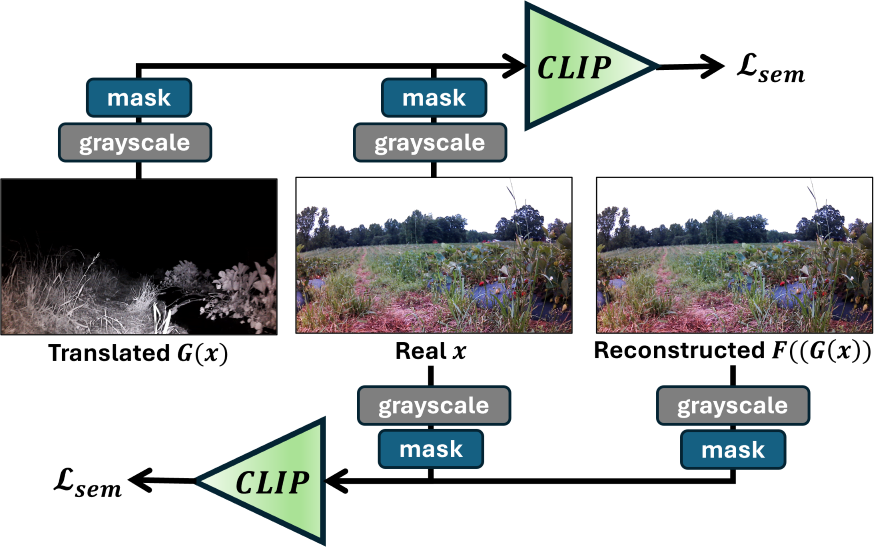}}
        \caption{
        Illustration of the data flow for computing (a)~the cycle consistency loss in $X \rightarrow Y \rightarrow X$ with generators~$G$ and~$F$ (Eq.~\ref{eq:cycle_consistency2}), and (b)~the semantic consistency loss between $x$ and its translated images using a pre-trained CLIP (Eq.~\ref{eq:semantic_consistency}).
            }
        \label{fig:architecture}
\end{figure*}

\section{Unsupervised Day-to-Night Cross-Modal Image Translation}

Our primary goal is to learn a mapping~$G: X \rightarrow Y$ that translates daytime RGB images~$x\in X$ into nighttime NIR images~$y\in Y$ \emph{without} requiring paired examples of the same scene. 
The semantic labels available in~$X$ are subsequently leveraged to train a semantic segmentation model~$\mathcal{S}$ using the translated images~$y$ (cf.~\autoref{fig:concept}), enabling robust nighttime operation.
As in~\citep{GLWPS20}, our framework builds upon CycleGAN~\citep{ZPIE17}, utilizing an adversarial setting and cycle consistency constraints. 
To ensure semantic preservation in translated imagery, we integrate pre-trained CLIP image encoders~\citep{RKHRGASAMC21}.
``Visibility'' masks are also introduced to simulate the restricted effective sensing range of NIR night-vision cameras. 
We denote the daytime and nighttime data distributions as $x \sim p_X$ and $y \sim p_Y$, respectively. 


\begin{figure}[t]\centering
    \subfloat[]{\label{fig:farm1_day}%
    \includegraphics[width=.45\linewidth]{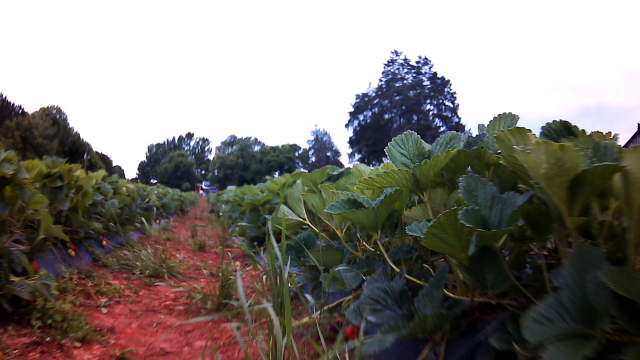}}
    \quad
    \subfloat[]{\label{fig:farm1_day_label}%
    \includegraphics[width=.45\linewidth]{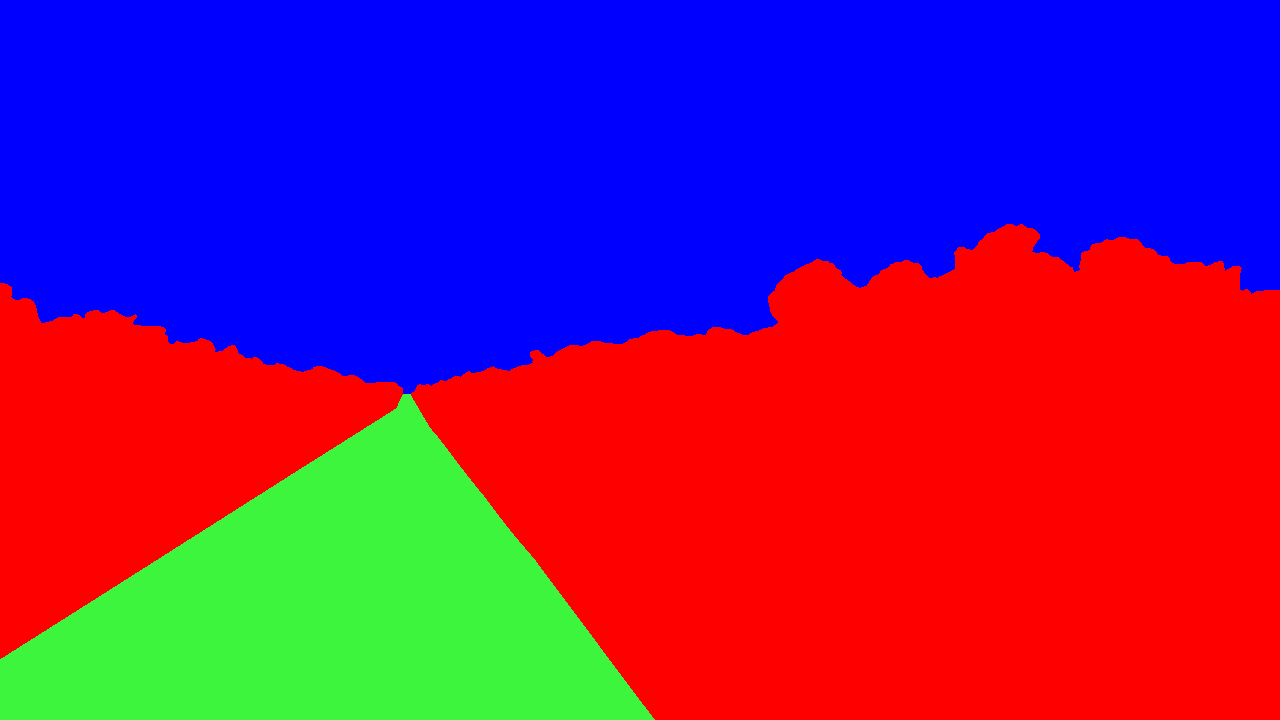}}\\
    \subfloat[]{\label{fig:farm2_day}%
    \includegraphics[width=.45\linewidth]{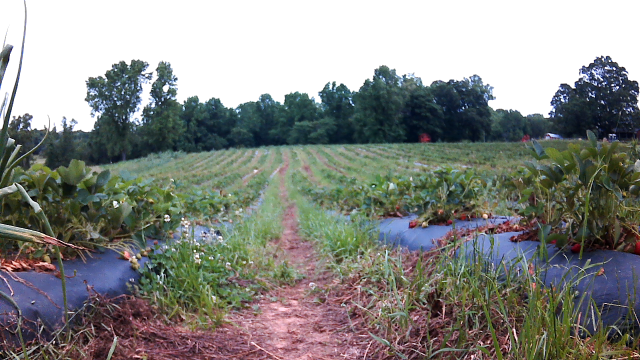}}
    \quad
    \subfloat[]{\label{fig:farm2_day_label}%
    \includegraphics[width=.45\linewidth]{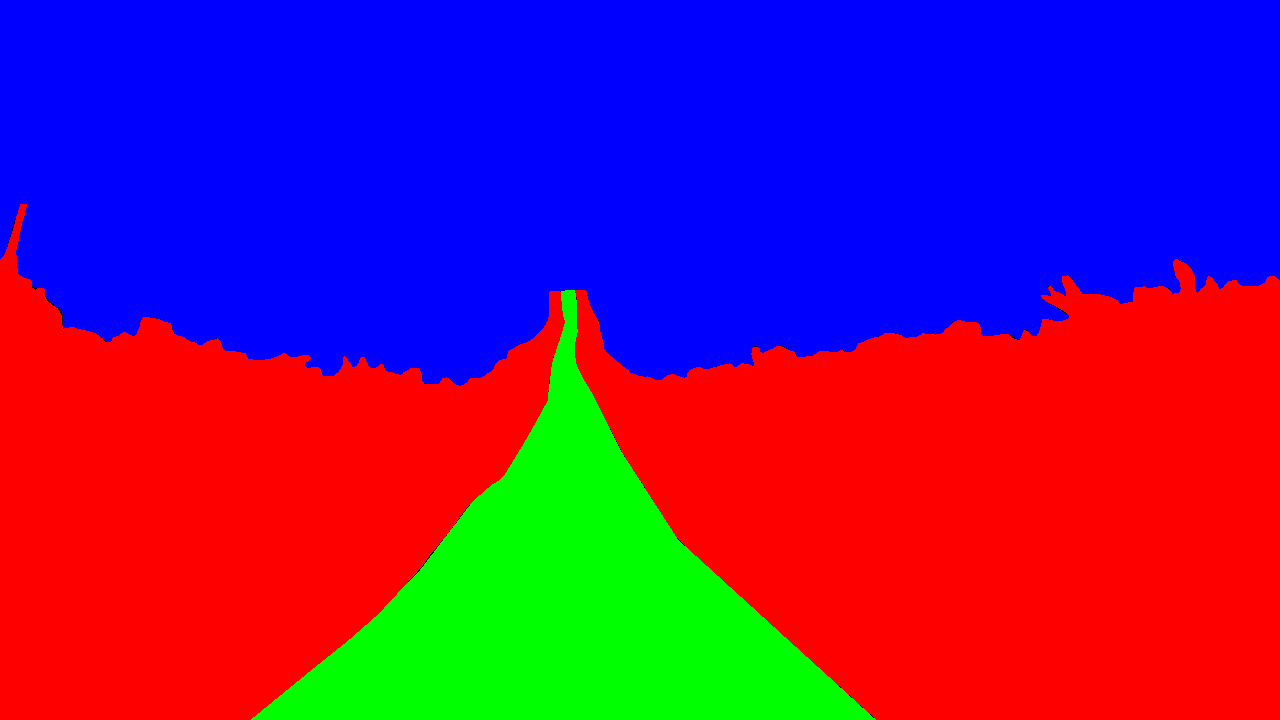}}\\
    \subfloat[]{\label{fig:farm2_night}%
    \includegraphics[width=.45\linewidth]{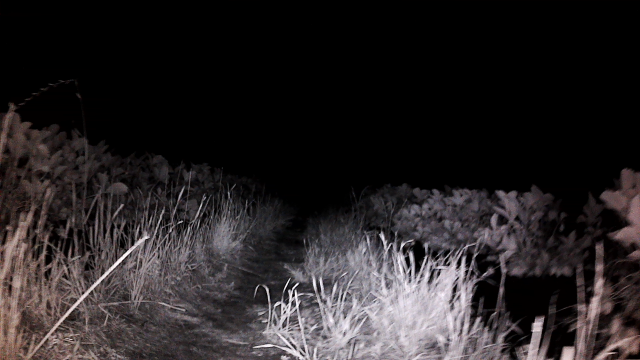}}
    \quad
    \subfloat[]{\label{fig:farm2_night_label}%
    \includegraphics[width=.45\linewidth]{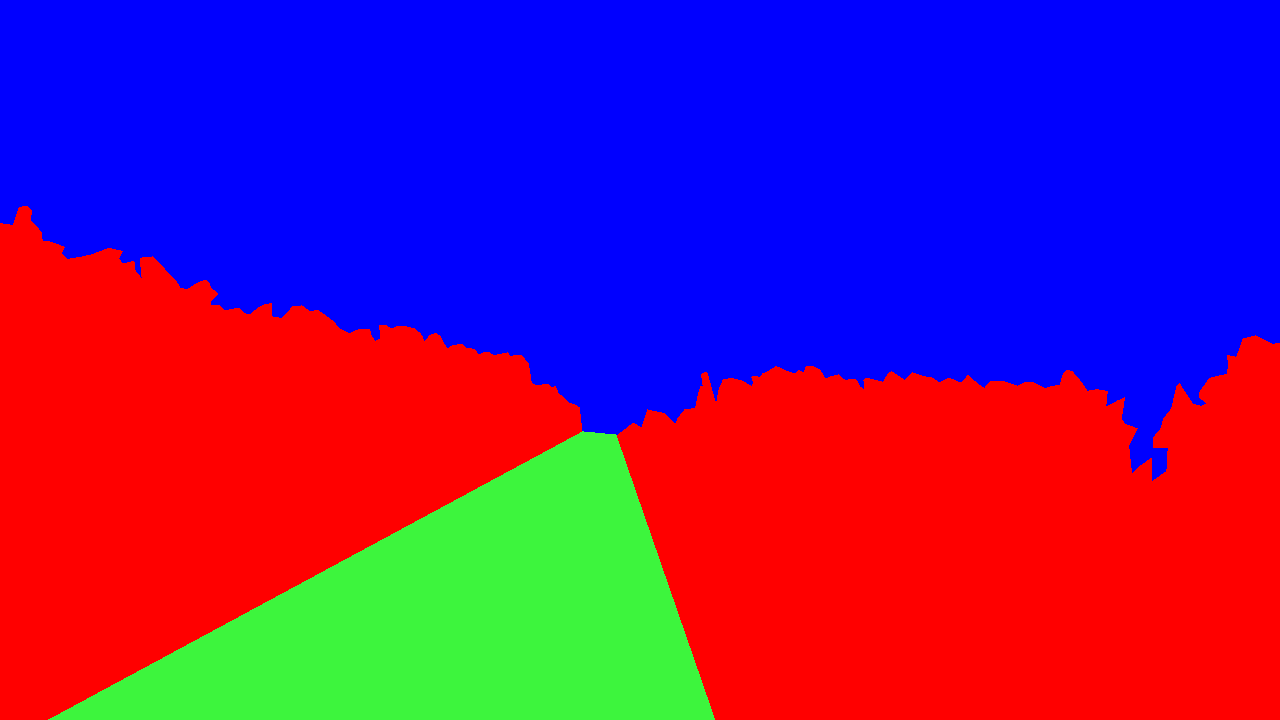}}\\
    \subfloat[]{\label{fig:carrot_farm_night}%
    \includegraphics[width=.45\linewidth]{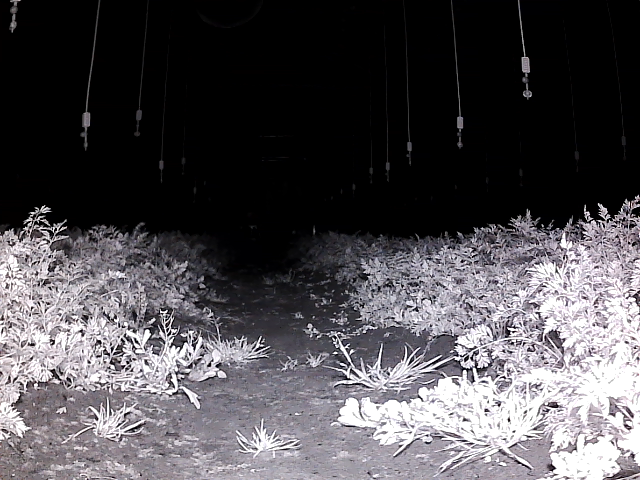}}
    \quad
    \subfloat[]{\label{fig:carrot_farm_night_label}%
    \includegraphics[width=.45\linewidth]{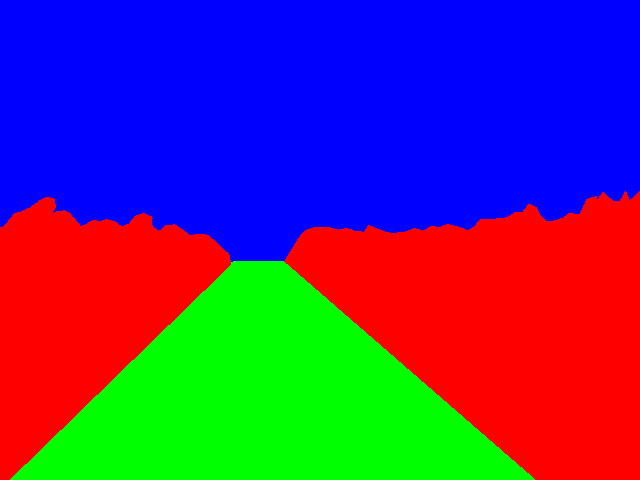}}
        \caption{
        (a)--(b) Daytime image and corresponding annotation from Strawberry Farm~A; 
        (c)--(f) Daytime and nighttime images with their respective annotations collected from Strawberry Farm~B; and (g)--(h) Nighttime image and corresponding annotation from the carrot field.  
            }
        \label{fig:data_exmaples}
\end{figure}

\subsection{Adversarial Training \& Masked Cycle Consistency}
\label{sec:adversarial_training}

In the adversarial setting~\citep{GPMXWOCB20}, a discriminator~$D_Y$ is trained to distinguish translated images~$G(x)$ from real nighttime images~$y$. 
Simultaneously, the generator~$G$ is optimized to produce realistic NIR images to deceive~$D_Y$. 
Inspired by Pix2Pix~\citep{IZZE17}, the discriminator~$D_Y$ and generator~$G$ are trained to minimize the following losses: 
%
%
\begin{equation}
    \begin{split}
    \mathcal{L}_{D_Y} = \mathbb{E}_{y \sim p_Y} \Big[ \frac{1}{HW}\|D_Y(y) - \mathbf{1}\|^2_F \Big]  \\
    + \mathbb{E}_{x \sim p_X} \Big[ \frac{1}{HW}\|D_Y(G(x))\|^2_F \Big],
    \end{split}
\label{eq:adversarial1}
\end{equation}
and
\begin{equation}
    \mathcal{L}_{G} = \mathbb{E}_{x \sim p_X} \Big[ \frac{1}{HW}\|D_Y(G(x)) - \mathbf{1}\|^2_F \Big],
\label{eq:adversarial2}
\end{equation}
where $D_Y(\cdot) \in \mathbb{R}^{H \times W}$ denotes the patch-level output from the discriminator,
$\mathbf{1}$ represents an $H \times W$ matrix of ones,
and $\|\cdot\|_F$ denotes the Frobenius norm.
To be specific, the discriminator~$D_Y$ evaluates realism over $HW$ subpatches, assigning values close to one for realistic patches and close to zero for fake ones, while the generator~$G$ aims to produce images that fool the discriminator into classifying generated patches as real. 
Following LSGAN~\citep{MLXLWP17}, we also adopt the least-squares (Frobenius norm-based) adversarial loss to alleviate the gradient vanishing problem during training.

To add a cycle consistency constraint, a secondary generator~$F: Y \rightarrow X$ and discriminator~$D_X$ are also trained in parallel to handle the inverse mapping (night to day),
using adversarial losses $\mathcal{L}_{D_X}$ and $\mathcal{L}_{F}$ analogous to Eq.\ref{eq:adversarial1} and~\ref{eq:adversarial2}, respectively. 
Both generators~$G$ and~$F$ are then trained to ensure an image translated through either~$X \rightarrow Y \rightarrow X$ (cf.~\autoref{fig:cycle1}) or~$Y \rightarrow X \rightarrow Y$ remains similar to its original input. 
To achieve this, $G$ and $F$ are optimized to minimize the cycle consistency loss~\citep{ZPIE17}: 
\begin{equation}
        \mathcal{L}_{cyc} = \mathbb{E}_{x\sim p_X}[\| F(G(x)) - x\|_1] 
        + \mathbb{E}_{y\sim p_Y}[\| G(F(y)) - y\|_1].
\end{equation}

As shown in~\autoref{fig:data_exmaples}, however, NIR night-vision sensors operating near an $850$~nm wavelength provide limited visibility, depending on the intensity and beam divergence of the infrared illumination~\citep{KB25}.
Consequently, distance regions in NIR nighttime images may lack sufficient information to reconstruct a full daytime scene. 
To account for this, we apply binary ``visibility'' masks~($m$ and $m'$) to the cycle consistency loss, ignoring pixels lying outside the sensor's reliable effective range in nighttime images. 
Specifically, the visibility mask~$m$ is created based on translated nighttime images~$G(x)$, while another mask~$m'$ is based on real nighttime images~$y$.
%
The resulting masked cycle consistency loss is defined as:
%
\begin{equation}
    \begin{split}
        \mathcal{L'}_{cyc} = \mathbb{E}_{x\sim p_X}[\| m \odot F(G(x)) - m \odot x\|_1] \\
        + \mathbb{E}_{y\sim p_Y}[\| m'\odot G(F(y)) - y\|_1],
    \end{split}
    \label{eq:cycle_consistency2}
\end{equation}
where $\odot$ denotes element-wise multiplication. 
Conceptually, this approach is similar to Mask CycleGAN~\citep{W22}. 
However, our objective is not to stitch a partially style-transferred image back into the original image. 

\subsubsection{Visibility Mask Generation}
\label{sec:visibility_mask}

Each visibility mask ($m, m' \in \{0,1\}^{H'\times W'}$) is a binary matrix with the same spatial dimensions as the corresponding image to which it is applied, where $H'$ and $W'$ denote the image height and width, respectively. 
To simulate the limited effective range of the NIR illumination, we set the ``visible'' region based on pixel intensity. 

Given a nighttime image (either a translated image~$G(x)$ for~$m$ or a real image~$y$ for~$m'$), we apply a heuristic algorithm that scans each column~$c_i$ to identify the uppermost row---i.e.,~the minimum row index~$r_j$---where the intensity exceeds a predefined threshold~$\tau$, representing the boundary of the illuminated, visible region.  
We then define the mask elements~$b(r, c_i)$ to be~$0$ if $r<r_j$ and~$1$ otherwise. 
This process is repeated for all columns to generate the final mask. 

Though the estimated mask may be imperfect during early training epochs when the translated image~$G(x)$ is still unstable, we observe that the inferred visibility boundary becomes gradually more realistic as adversarial training converges.
By incorporating this spatial constraint into the loss function, optimization is focused on regions where reliable NIR observations are available, \emph{avoiding} penalization for distant areas where the night-vision sensor lacks valid signals.
\Autoref{sec:conclusion} also discusses alternative designs.



\subsection{Semantic Consistency Enhancement}

To preserve the underlying semantic structure during translations, we utilize a fixed, pre-trained CLIP image encoder~$\mathcal{E}$~\citep{RKHRGASAMC21}. 
As illustrated in~\autoref{fig:clip1}, a cosine-similarity loss is computed in the shared CLIP embedding space to encourage the generators~$G$ and~$F$ to create translated images with semantic consistency with the original real images~$x$ or~$y$.
In this loss calculation, specifically, each input image is first converted to grayscale to reduce the influence of color information.  
The generators~$F$ and~$G$ are then optimized to minimize the semantic consistency loss below:
\begin{equation}
\begin{split}
    \mathcal{L}_{sem} = 
    &\quad \mathbb{E}_{x\sim p_X} \Bigl[1 - \cos_\mathcal{E} \Bigl( m \odot G(x), m \odot x \Bigr) \Bigr] \\
    &\quad + \mathbb{E}_{x\sim p_X} \Bigl[1 - \cos_\mathcal{E}\Bigl( m \odot F(G(x)), m \odot x \Bigr) \Bigr] \\
    &\quad + \mathbb{E}_{y\sim p_Y} \Bigl[1 - \cos_\mathcal{E}\Bigl( m' \odot F(y), y \Bigr) \Bigr] \\
    &\quad + \mathbb{E}_{y\sim p_Y} \Bigl[1 - \cos_\mathcal{E}\Bigl( m'\odot G(F(y)), y \Bigr) \Bigr],\\
\end{split}
\label{eq:semantic_consistency}
\end{equation}
where $\cos_\mathcal{E} (a, b)$ represents the cosine similarity between the embeddings~$\mathcal{E}(a')$ and $\mathcal{E}(b')$, where $a'$ and $b'$ are the grayscale versions of $a$ and $b$, respectively. 
Note here that the visibility binary masks ($m$ and $m'$) are also applied. 

Finally, the ultimate objective function is:
\begin{equation}
    \begin{split}
        \mathcal{L} = \lambda_{GAN}(\mathcal{L}_{D_Y} + \mathcal{L}_{G} + \mathcal{L}_{D_X} + \mathcal{L}_{F}) \\
        + \lambda_{cyc}\mathcal{L'}_{cyc} 
        + \lambda_{sem}\mathcal{L}_{sem} 
        + \lambda_{id}\mathcal{L}_{id}, 
    \end{split}
\end{equation}
where $\mathcal{L}_{id} = \mathbb{E}_{y\sim p_Y}\|G(y) - y\|_1 + \mathbb{E}_{x\sim p_X}\|F(x) - x\|_1$---i.e.,~the identity mapping loss to regularize the generators by ensuring target-domain samples are preserved when passed through their respective generators~\citep{ZPIE17}.
The hyperparameters~$\lambda$ control the relative weight of each loss component.

\section{Experiments}
\label{sec:experiments}

\subsection{AgriNight Dataset}
\label{sec:agrinight_dataset}

As summarized in~\autoref{tab:dataset_stat}, the AgriNight dataset contains $428$~daytime RGB and $549$~nighttime NIR images with pixel-wise semantic labels across three classes: \texttt{Traversable} (inter-row space), \texttt{Non-Traversable} (crop rows and plant regions), and \texttt{Other}. 
The pixel distribution for each class is reported in~\autoref{tab:dataset_stat2}, where \texttt{Traversable} regions occupy the smallest proportion of pixels.
Note that due to limited visibility during nighttime (cf.~\autoref{fig:data_exmaples}), the proportion of the \texttt{Other} class increases, while the proportions of the \texttt{Traversable} and \texttt{Non-Traversable} classes decrease.

Data were collected through multiple sessions conducted during daytime and nighttime conditions across two large-scale commercial strawberry farms in Georgia, USA, and one carrot field located at the KSU Field Station---a research farm operated by Kennesaw State University in the same state. 
\Autoref{fig:data_exmaples} shows example images and labels.

\subsubsection{Strawberry Fields}

In Strawberry Farm~A, both daytime and nighttime images were collected from five crop rows, each approximately $40$~meters in length. 
In Strawberry Farm~B, data were sampled from nine crop rows, each approximately $70$~m long and $0.8$~m wide.
As shown in~\autoref{fig:farm1_day} and~\ref{fig:farm2_day}, the fields in Farm~A were predominantly flat, while crop rows in Farm~B exhibited a shallow longitudinal depression centered along each row.
This structural difference introduced visual \emph{variability} to the dataset, which may improve model generalization to novel field environments.
In both fields, natural weeds were frequently observed within crop rows and along inter-row regions. 
All strawberry fields were cultivated using plastic-mulched raised beds, a common plasticulture system in commercial production.

During each session, a Rover Robotics $4$WD Rover Zero~$3$ platform was manually operated, with the operator relying solely on natural moonlight for visibility. 
To ensure diverse visual observations, the robot was driven along crop rows using both straight and zigzag trajectories. 
The platform was equipped with an Arducam night-vision camera featuring six integrated $850$~nm NIR LEDs and a switchable infrared-cut filter that automatically toggles based on light conditions. 
The camera was mounted approximately $33$~cm above the ground at the front of the robot to capture both daytime and nighttime images at a resolution of $1280 \times 720$~pixels.

A total of $331$~daytime and $370$~nighttime images were collected over two days and two nights in mid-May during the harvesting season, when strawberry plants were mature, densely foliated, and bearing fruit. 
Daytime sessions were conducted under clear weather conditions. 
Nighttime data collection was performed after sunset (\mbox{9PM} to midnight) to capture realistic low-light field environments.

\subsubsection{Carrot Fields}

The carrot field was located in a greenhouse environment, where six crop rows were covered, each extending approximately $24$~m in length and $0.6$~m in width.  
As shown in~\autoref{tab:dataset_stat}, a total of $97$~daytime and $179$~nighttime images were collected in late February, approximately two months after planting. 
At the time of data collection, each carrot plant was approximately $19$~cm tall.

We utilized a similar robot and camera setting to that used in the strawberry fields. 
To better capture these relatively small plants, however, the camera was mounted at a lower height, approximately $22$~cm above ground level, along with an Andoer portable LED, featuring $49$~NIR lamps, installed above the camera to provide additional active illumination during nighttime sessions.
Both daytime and nighttime images were captured at a resolution of $640 \times 480$~pixels. 



\begin{figure}[t]\centering
    \subfloat[]{\label{fig:good1}%
    \includegraphics[width=.45\linewidth]{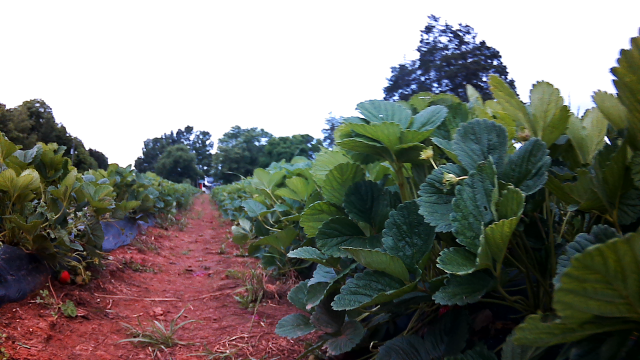}}
    \quad
    \subfloat[]{\label{fig:bad1}%
    \includegraphics[width=.45\linewidth]{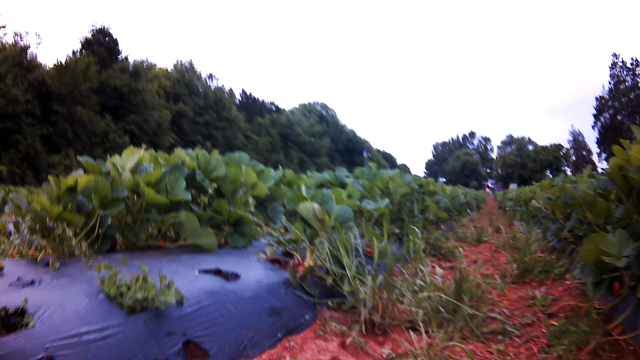}}\\
    \subfloat[]{\label{fig:good2}%
    \includegraphics[width=.45\linewidth]{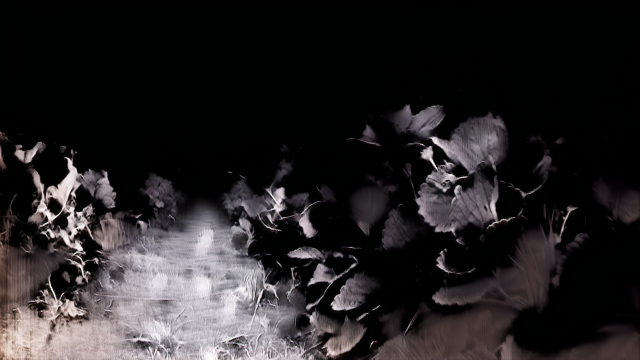}}
    \quad
    \subfloat[]{\label{fig:bad2}%
    \includegraphics[width=.45\linewidth]{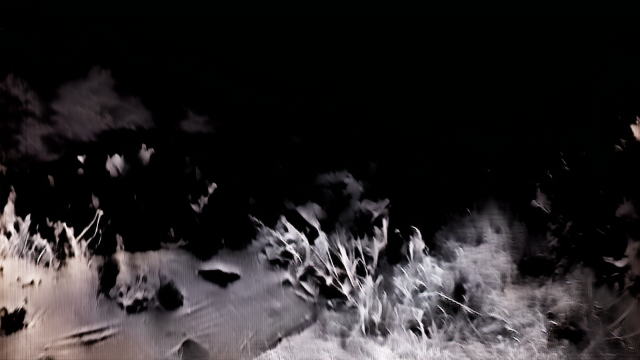}}\\
    \subfloat[]{\label{fig:good3}%
    \includegraphics[width=.45\linewidth]{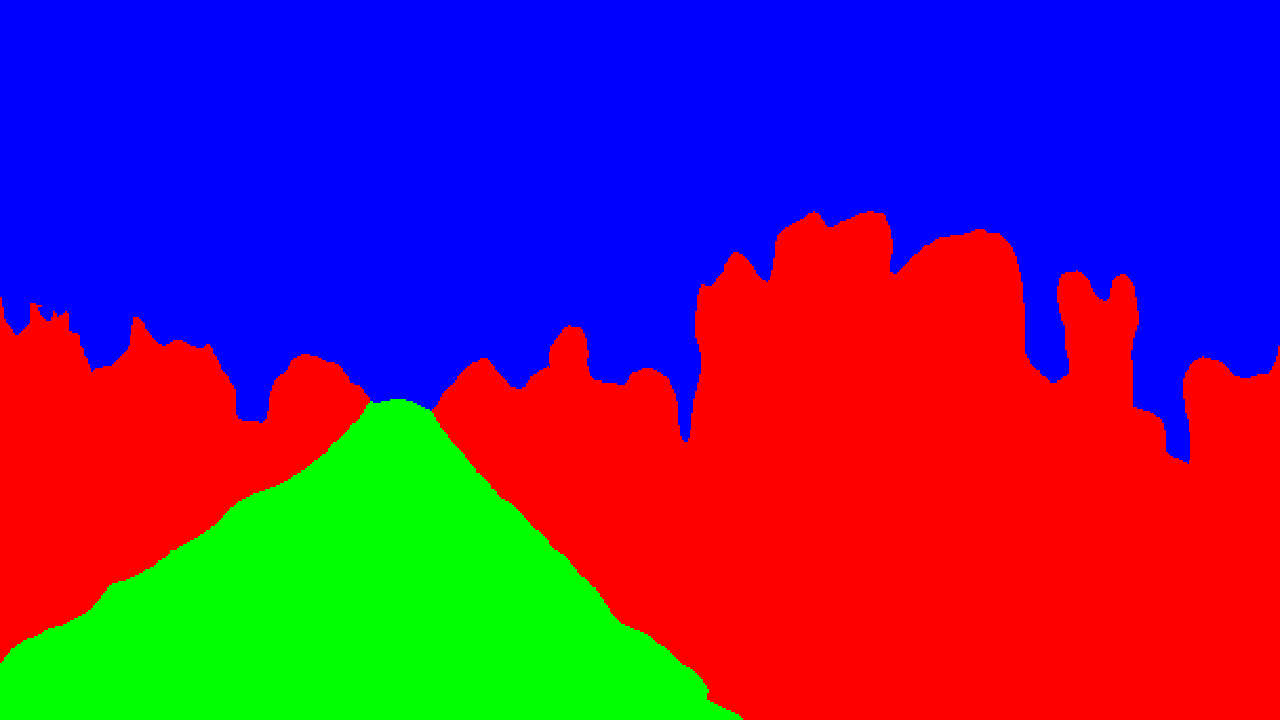}}
    \quad
    \subfloat[]{\label{fig:bad3}%
    \includegraphics[width=.45\linewidth]{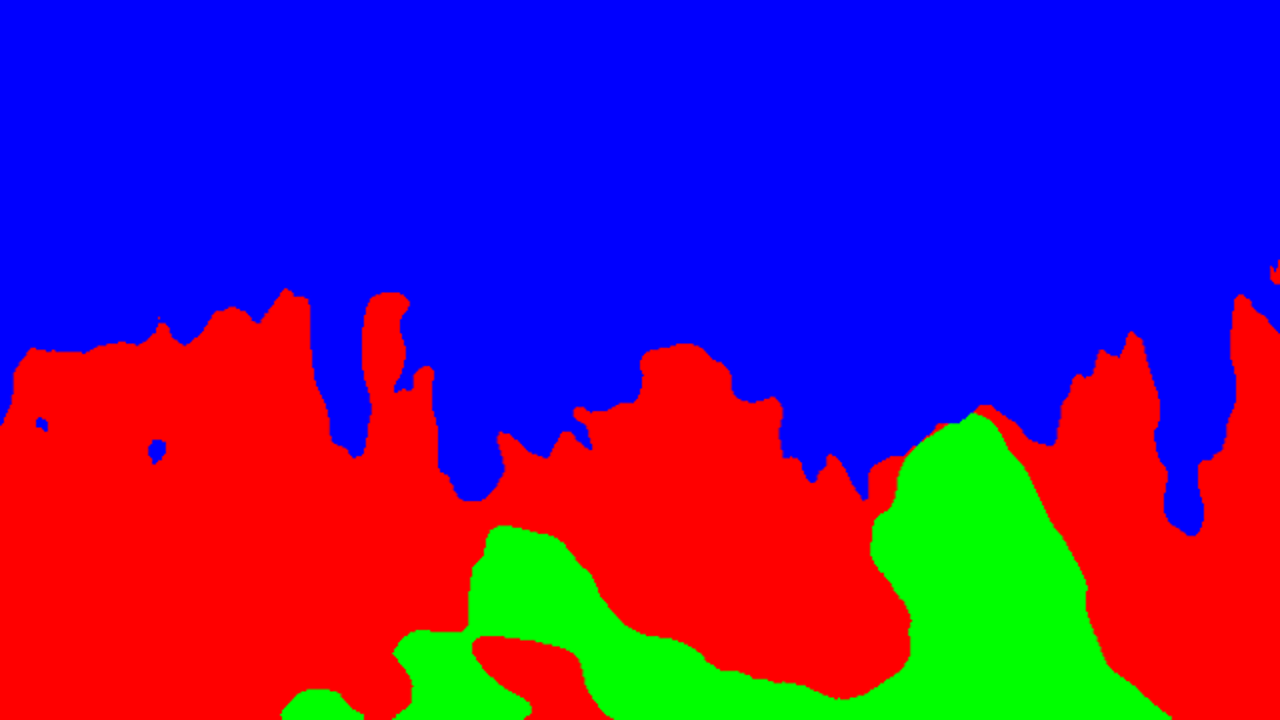}} \\
    %
        \caption{
        Given the daytime images in (a)--(b), the proposed method synthesized a high-quality nighttime image in (c) and a low-quality one in (d), resulting in accurate segmentation in (e) and erroneous segmentation in (f), respectively. 
        More examples are available in the code repository.
            }
        \label{fig:prediction_examples}
\end{figure}

\subsection{Implementation Details}
\label{sec:implementation_details}



Inspired by~\citep{GLWPS20}, each of our generators (i.e.,~$G$ and $F$) follows an encoder-transformer-decoder architecture.
To process an input image of size~$256 \times 256$, the encoder utilizes a $2$D convolutional layer with $64$~filters of size~$7\times 7$, followed by a second $2$D convolutional layer with $128$~filters of size~$3 \times 3$ and stride~$2$ for spatial downsampling.
The transformer module consists of $16$ sequential residual blocks, each equipped with instance normalization layers instead of batch normalization and without additional spatial downsampling.
The decoder employs a $2$D transposed convolution layer with $64$ filters of size $3\times 3$ kernels and stride~$2$ to restore the original spatial resolution through learnable upsampling, followed by a $2$D convolutional layer with three~$7 \times 7$ filters to produce a three-channel output image.

Each discriminator ($D_X$ and $D_Y$) comprises five $2$D convolutional layers, with the first four using $64, 128, 256$ and $512$ filters of size~$4 \times 4$ with stride~$2$, respectively. 
Each layer applies instance normalization.
The final layer uses a $4 \times 4$ convolution to produce a single-channel $16 \times 16$ subpatch prediction map, which is used for the computation of $\mathcal{L}_{GAN}$. 
During training, standard image augmentation methods such as random cropping and horizontal flip were performed with a probability of $25\%$. 

To utilize the embedding space from a pretrained CLIP model, we utilized the ViT-B/$32$ Transformer image encoder\footnote{https://huggingface.co/openai/clip-vit-base-patch32}. 
Additionally, a semantic segmentation model was designed with a ResNet-$18$ backbone~\citep{TAL16} as the encoder, followed by a decoder with symmetric upsampling layers and U-Net–style skip connections~\citep{GLWPS20}.
The model performs pixel-wise classification over three semantic classes: \texttt{Traversable}, \texttt{Non-Traversable}, and \texttt{Other}. 
Training is conducted using the standard cross-entropy loss for multi-class segmentation~\citep{GKCSWM22}.
The model was trained for $100$~epochs, and we selected the checkpoint that achieves the highest mean Intersection-over-Union~(mIoU) for evaluation.
During training, data augmentation methods, including random cropping, horizontal flipping, Gaussian blurring, and color jittering, were applied with a probability of $25\sim50\%$. 

All training was conducted with a single NVIDIA RTX A5000 GPU ($24$~GB VRAM). 
Our image translation model was trained for $500$~epochs with a learning rate of $10^{-5}$, and the checkpoint achieving the lowest FID score---designed to assess overall image quality~\citep{PPNZY24}---was selected for evaluation. 
The loss weights were set to $\lambda_{cyc}=10$, $\lambda_{GAN}=1$, $\lambda_{id}=5$, and $\lambda_{sem}=1$ based on empirical validation.



\begin{table}[!t]
    \small
    \centering
    
    \begin{tabular}{|C{20mm}||C{16mm}|C{16mm}|C{16mm}|}
        \hline
         & \textbf{\# Day} & \textbf{\# Night} & \textbf{\# Rows} \\
        \hline
        \hline
        \textbf{Total} & {428} & {549} & {20} \\
        \hline
        \textbf{Strawberry A} & 181 & 185 & 5 \\
        \hline
        \textbf{Strawberry B} & 150 & 185 & 9 \\
        \hline
        \textbf{Carrot} & 97 & 179 & 6 \\
        \hline
    \end{tabular}
    \caption{Summary of collected daytime and nighttime images and the number of crop rows covered in each farm. 
    }
    \label{tab:dataset_stat}
\end{table}

\begin{table}[!t]
    \small
    \centering
    
    \begin{tabular}{|C{13mm}||C{18mm}|C{24mm}|C{18mm}|}
        \hline
         & \textbf{Traversable} & \textbf{Non-Traversable} & \textbf{Other} \\
        \hline
        \hline
        \textbf{Day} & $\mathbf{15.5\%}$ & $\mathbf{37.8\%}$ & $46.7\%$ \\
        \hline
        \textbf{Night} & $12.2\%$ & $33.9\%$ & $\mathbf{53.9\%}$\\
        \hline
    \end{tabular}
    \caption{Classs-wise pixel distribution for daytime and nighttime images in the AgriNight dataset.}
    \label{tab:dataset_stat2}
\end{table}

\subsection{Evaluation Protocols}
\label{sec:evaluation_protocols}

For evaluation, we focus on the downstream nighttime semantic segmentation task, which is a critical component for achieving reliable nighttime visual navigation. 
As shown in~\autoref{fig:concept}, the segmentation model is trained exclusively on translated nighttime images and masked daytime annotations, and validated on real nighttime images. 
In particular, the visibility mask~$m$ (cf.~\autoref{sec:visibility_mask}) is applied to the daytime annotations to emulate reduced nighttime visibility.
Following prior work~\citep{GKCSWM22, WERHK19}, we report mIoU and average pixel-wise accuracy as the primary evaluation metrics. 
In addition, we compute precision and recall~\citep{MWGLBPS23} for each class.

We primarily utilize data from the Strawberry Fields dataset (\autoref{sec:agrinight_dataset}) due to its relatively large volume, while reserving the Carrot Field dataset for physical robot experiments. 
Specifically, the nighttime NIR~images from the Strawberry Fields dataset are organized into three predefined splits, each containing mutually exclusive \texttt{Train} ($247$~images) and \texttt{Val} ($123$~images) subsets, enabling $3$-fold cross-validation.
In each fold, all daytime images together with the nighttime images from \texttt{Train} are used for unsupervised training of the day-to-night image translation model. 
The translated nighttime images from all daytime images and their masked semantic labels are then used to train the downstream segmentation model (\autoref{fig:concept}). 
The trained segmentation model is subsequently evaluated on \texttt{Val}. 

To prevent temporal leakage between \texttt{Train} and \texttt{Val} subsets, image frames collected from each farm are first chronologically ordered according to sampling time and then partitioned into three equally sized sequential segments. 
The first split then assigns the first temporal segment from each farm to the \texttt{Val} set, while the remaining two segments are used for \texttt{Train}. 
The other splits are constructed by rotating the validation segment accordingly.
As a result, we report the average performance across the three distinct \texttt{Val} sets.



\begin{table*}[t]
    \small
    \centering
    \begin{tabular}{|C{10mm}|C{20mm}||C{13mm}|C{13mm}|C{8mm}|C{8mm}|C{8mm}|C{8mm}|C{8mm}|C{8mm}|C{8mm}|C{8mm}|C{8mm}|}
    \hline
     \multicolumn{2}{|c||}{} & \multirow{2}{*}{mIoU} & \multirow{2}{*}{\makecell{Pixel\\Accuracy}} & \multicolumn{3}{c|}{Traversable} \normalsize & \multicolumn{3}{c|}{Non-Traversable} \normalsize & \multicolumn{3}{c|}{Other} \normalsize\\
        \cline{5-13}
     \multicolumn{2}{|c||}{} & & & Prec. & Rec. & F1 & Prec. & Rec. & F1 & Prec. & Rec. & F1 \\
    \hline
    \hline
    \multicolumn{2}{|c||}{UoB~\cite{GLWPS20}} & $.784$\newline\tiny$\pm.017$  & $.905$\newline\tiny$\pm.007$ & $.745$\newline\tiny$\pm.025$ & $\mathbf{.868}$\newline\tiny$\pm.03$ & $.801$\newline\tiny$\pm.025$ & $.901$\newline\tiny$\pm.016$ & $.804$\newline\tiny$\pm.016$ & $.850$\newline\tiny$\pm.011$ & $.962$\newline\tiny$\pm.002$ & $.979$\newline\tiny$\pm.006$ & $.971$\newline\tiny$\pm.002$ \\
    \hline
    \multicolumn{2}{|c||}{SSL-RGB2IR~\cite{SSAS24}} & $.754$\newline\tiny$\pm.038$  & $.890$\newline\tiny$\pm.021$ & $.703$\newline\tiny$\pm.089$ & $.825$\newline\tiny$\pm.019$ & $.756$\newline\tiny$\pm.048$ & $.869$\newline\tiny$\pm.018$ & $.786$\newline\tiny$\pm.057$ & $.824$\newline\tiny$\pm.038$ & $.973$\newline\tiny$\pm.01$ & $.976$\newline\tiny$\pm.004$ & $.974$\newline\tiny$\pm.004$\\
    \hline
    \multicolumn{2}{|c||}{CycleGAN-Turbo~\cite{PPNZY24}} & $.745$\newline\tiny$\pm.04$  & $.892$\newline\tiny$\pm.018$ & $.799$\newline\tiny$\pm.025$ & $.649$\newline\tiny$\pm.102$ & $.713$\newline\tiny$\pm.072$ & $.816$\newline\tiny$\pm.048$ & $\mathbf{.879}$\newline\tiny$\pm.026$ & $.845$\newline\tiny$\pm.020$ & $.969$\newline\tiny$\pm.013$ & $.973$\newline\tiny$\pm.015$ & $.971$\newline\tiny$\pm.002$\\
    \hline
    \multirow{2}{*}{Ours} & $9$~Res.~Blocks & $.814$\newline\tiny$\pm.027$  & $.921$\newline\tiny$\pm.013$ & $.813$\newline\tiny$\pm.037$ & $.838$\newline\tiny$\pm.020$ & $.825$\newline\tiny$\pm.028$ & $.887$\newline\tiny$\pm.019$ & $.873$\newline\tiny$\pm.025$ & $\mathbf{.880}$\newline\tiny$\pm.022$ & $\mathbf{.976}$\newline\tiny$\pm.002$ & $.976$\newline\tiny$\pm.005$ & $\mathbf{.976}$\newline\tiny$\pm.003$ \\
    \cline{2-13}
     & $16$~Res.~Blocks & $\mathbf{.839}$\newline\tiny$\pm.011$  & $\mathbf{.932}$\newline\tiny$\pm.005$ & $\mathbf{.841}$\newline\tiny$\pm.018$ & $\mathbf{.868}$\newline\tiny$\pm.015$ & $\mathbf{.854}$\newline\tiny$\pm.013$ & $\mathbf{.916}$\newline\tiny$\pm.024$ & $.842$\newline\tiny$\pm.039$ & $.877$\newline\tiny$\pm.020$ & $.971$\newline\tiny$\pm.004$ & $\mathbf{.978}$\newline\tiny$\pm.013$& $.974$\newline\tiny$\pm.005$ \\
     \hline\hline
     \multicolumn{2}{|c||}{Real Nighttime} & $.922$\newline\tiny$\pm.003$  & $.969$\newline\tiny$\pm.002$ & $.931$\newline\tiny$\pm.012$ & $.942$\newline\tiny$\pm.015$ & $.936$\newline\tiny$\pm.002$ & $.954$\newline\tiny$\pm.009$ & $.952$\newline\tiny$\pm.009$ & $.953$\newline\tiny$\pm.003$ & $.989$\newline\tiny$\pm.001$ & $.987$\newline\tiny$\pm.003$ & $.988$\newline\tiny$\pm.001$\\
    \hline
    \end{tabular}
    \caption{Comparative evaluation results reported with the mean and standard deviation over three-fold cross-validation on the AgriNight Strawberry Field dataset. Bold values indicate the best result per metric, excluding Real Nighttime---the upper-bound peformance obtained from training with annotated real nighttime images.}
    \label{tab:comparison}
\end{table*}

\subsection{Comparative Evaluations}
\label{sec:comparative_eval}

The proposed framework is compared against representative image-to-image translation and domain adaptation models discussed in~\autoref{sec:related_work}:
\begin{itemize}
     \item \textbf{UoB}~\cite{GLWPS20}: An unsupervised domain adaptation framework developed at the University of Bonn for crop, weed, and soil semantic segmentation in unseen fields. 
     \item \textbf{SSL-RGB2IR}~\cite{SSAS24}: A RGB-to-TIR image translation framework designed to support object detection and semantic segmentation in urban environments. 
    \item \textbf{CycleGAN-Turbo}~\cite{PPNZY24}: A CycleGAN-style framework that leverages a \emph{pretrained} Stable Diffusion-Turbo~\cite{SLBR24} backbone as the generator for image translation.
    \item \textbf{Real Nighttime}: The segmentation model is trained using annotated real nighttime NIR images in~\texttt{Train} to demonstrate the achievable upper-bound performance.
\end{itemize}

Both SSL-RGB2IR and CycleGAN-Turbo were trained and validated using their official implementations with task-specific modifications. 
In contrast, UoB was reimplemented based on the published description due to the absence of publicly released code. 
The primary modification in our implementation is the use of nine residual blocks, instead of the six, to increase model capacity for cross-modal translation.


For a fair comparison, the hyperparameters of each baseline model (e.g.,~learning rates and loss weights) were tuned to minimize the FID score~\citep{PPNZY24}. 
All adopted hyperparameters, as well as the text prompts used for CycleGAN-Turbo, are available in the code repository.
In addition, the FID score was used to select the best-performing checkpoint during training. 
Similarly, for the downstream segmentation task, the checkpoint achieving the highest mIoU on translated nighttime images was selected for evaluation.
The final configurations are documented in our code repository.

\subsubsection{Result Analysis}

\Autoref{tab:comparison} presents the three-fold cross validation results of the semantic segmentation task evaluated on real nighttime data from the AgriNight Strawberry Fields dataset. 
The proposed model with $16$ residual blocks achieves the highest performance across most evaluation metrics,
compared to all baseline methods. 
Moreover, it exhibits lower standard deviations across different \texttt{Val} sets, indicating more consistent performance.
This result suggests that the translated nighttime images used for training effectively preserve structural and visual characteristics of real nighttime NIR data. 
Although the model achieves approximately $9\%$ lower mIoU ($4\%$ lower accuracy) than the upper-bound scenario trained directly on annotated real nighttime images, the results remain impressive given that training relied solely on translated images derived from daytime RGB inputs.

In particular, all methods tend to struggle with the \texttt{Traversable} class, likely due to its relatively small pixel proportion in the dataset (cf.~\autoref{tab:dataset_stat2}), which limits effective learning. 
Additionally, visual ambiguities caused by weeds, debris, and irregular soil textures may further confuse the segmentation models.
However, the performance gain of our proposed model is particularly pronounced for this class, which is critical for reliable robot navigation.

Based on mIoU and pixel-wise accuracy, UoB, which also employs a dedicated semantic consistency loss, produces higher-quality NIR translations than SSL-RGB2IR and CycleGAN-Turbo. 
Yet, it remains inferior to our model using nine residual blocks---the same number used in UoB. 
This indicates that integrating a pretrained CLIP model, rather than jointly training domain-specific segmentation models, contributes to producing higher-fidelity NIR translations.

Although CycleGAN-Turbo leverages pretrained Stable Diffusion Turbo, its poor performance and high variance imply that transferred knowledge from training on large-scale data does not generalize well to agricultural NIR imagery.

\subsection{Qualitative Analysis}

Given the daytime scene in~\autoref{fig:good1}, the translated image (\autoref{fig:good2}) presents a well-structured nighttime appearance, leading to precise semantic segmentation with clear boundaries of the \texttt{Traversable} region (\autoref{fig:good3}). 
While soil textures and some nearby leaves are not rendered in detail, trees in the daytime background become invisible in~\autoref{fig:good2}. 


However, \autoref{fig:bad2} shows missing strawberry plants on the raised bed and artifacts in the left background, causing the segmentation model to misclassify weeds within the \texttt{Traversable} region as \texttt{Non-Traversable} plants and parts of the raised bed as \texttt{Traversable} in~\autoref{fig:bad3}.
Using a larger CLIP model might offer better semantic alignment.

\subsection{Ablation Study}
\label{sec:ablation_study}

\Autoref{tab:ablation} summarizes our ablation study on the impacts of two core components: visibility masking mechanism and CLIP-based semantic consistency constraint.
The na\"ive model without either component yields the worst performance. 
Using a CLIP results in a more substantial improvement than applying visibility masking alone, highlighting the importance of semantic alignment in cross-modal translation. 

Finally, combining both components achieves the best overall performance, validating the proposed design choice. 
The \emph{complementary} effects of NIR-specific visual regularization and semantic guidance demonstrate significant improvement in day-to-night cross-modal image translation. 


\subsection{Physical Field Testing}

The robot platform introduced in~\autoref{sec:agrinight_dataset} was later deployed in the same carrot field used to acquire the Carrot Field dataset.
The $16$~residual block-model, pre-trained on the Strawberry Fields data, was further fine-tuned using all Carrot Field data to learn the RGB-to-NIR translation. 
The robot was equipped with an NVIDIA Jetson Orin for onboard control, running the trained segmentation model in real time.

For simplicity, the robot employed a ``pure pursuit'' controller to follow paths generated from the segmentation output, keeping the robot centered within the \texttt{Traversable} region.
Our framework is, however, controller-agnostic and can be readily integrated with other controllers. 

During autonomous navigation at a speed of $0.2$~m/s, the onboard system processed $14$~frames per second.
Over a total travel distance of $96$~m, the robot ran \emph{without} any collisions.
This experiment highlights the generalizability of our approach, as testing was conducted in previously unseen crop rows, and plants showed increased heights ($25$~cm) and denser foliage than those in the training data. 
A visual demonstration is provided in the supplementary video.


\begin{table}[t]
    \small
    \centering
    \begin{tabular}{|C{9mm}|C{9mm}||C{9mm}|C{13mm}|C{8mm}|C{8mm}|C{8mm}|}
    \hline
     \multirow{2}{*}{Mask} & \multirow{2}{*}{CLIP} &  \multirow{2}{*}{mIoU} & \multirow{2}{*}{\makecell{Pixel\\Accuracy}} & \multicolumn{3}{c|}{Traversable} \normalsize  \normalsize\\
    \cline{4-7}
      & &  & & Prec. & Rec. & F1 \\
      \hline
      \hline
        & & $.746$\newline\tiny$\pm.022$  & $.887$\newline\tiny$\pm.011$ & $.690$\newline\tiny$\pm.037$ & $.823$\newline\tiny$\pm.058$ & $.749$\newline\tiny$\pm.036$ \\
      \hline
      \checkmark &  & $.749$\newline\tiny$\pm.029$  & $.888$\newline\tiny$\pm.016$ & $.700$\newline\tiny$\pm.033$ & $.814$\newline\tiny$\pm.027$ & $.752$\newline\tiny$\pm.027$ \\
      \hline
       & \checkmark & $.806$\newline\tiny$\pm.048$  & $.916$\newline\tiny$\pm.024$ & $.796$\newline\tiny$\pm.109$ & $.856$\newline\tiny$\pm.029$ & $.819$\newline\tiny$\pm.053$ \\
      \hline
      \checkmark & \checkmark & $\mathbf{.839}$\newline\tiny$\pm.011$  & $\mathbf{.932}$\newline\tiny$\pm.005$ & $\mathbf{.841}$\newline\tiny$\pm.018$ & $\mathbf{.868}$\newline\tiny$\pm.015$ & $\mathbf{.854}$\newline\tiny$\pm.013$ \\
      \hline
     \end{tabular}
         \caption{Ablation study results evaluating the impact of the masking mechanism and CLIP loss in the proposed framework. 
         Checkmarks (\checkmark) indicate enabled components.}
    \label{tab:ablation}
 \end{table}

\section{Conclusion, Limitations, \& Future Work}
\label{sec:conclusion}




An unsupervised RGB-to-NIR translation framework is proposed for nighttime agricultural navigation. 
Built upon CycleGAN~\citep{ZPIE17}, it leverages a pre-trained CLIP model~\citep{RKHRGASAMC21} for semantic consistency and a visibility mask for limited NIR illumination range. 
Segmentation models trained on translated images perform reliably on real nighttime data, demonstrating high-fidelity translation, and are deployed alongside a robot controller to enable autonomous nighttime navigation.
The AgriNight dataset, comprising daytime/nighttime images with pixel-wise semantic labels, is also introduced.

However, crop-row switching~\citep{DCG23}, which is essential for fully autonomous field robots, is not addressed here, and mask realism warrants further investigation. 
Future work may explore learning-based masking, depth estimation for improved physical consistency, and larger CLIP models.

\section*{Acknowledgment}

This work was partially supported by USDA-NIFA (2023-38640-39572) through the Southern SARE program (GS24-301) and by the Georgia Peanut Commission (KSU-1-26/26). 

\addtolength{\textheight}{-12cm}   









{\small
    \ifx\usenatbib\undefined%
	\bibliographystyle{IEEEtran}%
    \else%
    \bibliographystyle{IEEEtranN}%
    \fi
	\bibliography{references}

@article{SGMHAC24,
  title={Demonstrating CropFollow++: Robust Under-Canopy Navigation with Keypoints},
  author={Sivakumar, AN and Gasparino, MV and McGuire, M and Higuti, VAH and Akcal, MU and Chowdhary, G},
  journal={RSS},
  year={2024}
}

@article{DCWG24,
  title={Deep learning-based crop row detection for infield navigation of agri-robots},
  author={De Silva, Rajitha and Cielniak, Grzegorz and Wang, Gang and Gao, Junfeng},
  journal={Journal of field robotics},
  year={2024},
  publisher={Wiley Online Library}
}

@inproceedings{GLWPS20,
  title={Unsupervised domain adaptation for transferring plant classification systems to new field environments, crops, and robots},
  author={Gogoll, Dario and Lottes, Philipp and Weyler, Jan and Petrinic, Nik and Stachniss, Cyrill},
  booktitle={IROS},
  year={2020},
}

@inproceedings{ZPIE17,
  title={Unpaired image-to-image translation using cycle-consistent adversarial networks},
  author={Zhu, Jun-Yan and Park, Taesung and Isola, Phillip and Efros, Alexei A},
  booktitle={ICCV},
  year={2017}
}

@inproceedings{RKHRGASAMC21,
  title={Learning transferable visual models from natural language supervision},
  author={Radford, Alec and Kim, Jong Wook and Hallacy, Chris and Ramesh, Aditya and Goh, Gabriel and Agarwal, Sandhini and Sastry, Girish and Askell, Amanda and Mishkin, Pamela and Clark, Jack and others},
  booktitle={ICML},
  year={2021},
}

@article{LJCK23,
  title={Edge-guided multi-domain rgb-to-tir image translation for training vision tasks with challenging labels},
  author={Lee, Dong-Guw and Jeon, Myung-Hwan and Cho, Younggun and Kim, Ayoung},
  journal={arXiv},
  year={2023}
}

@inproceedings{SSAS24,
  title={{SSL-RGB2IR}: Semi-supervised rgb-to-ir image-to-image translation for enhancing visual task training in semantic segmentation and object detection},
  author={Sikdar, Aniruddh and Saadiyean, Qiranul and Anand, Prahlad and Sundaram, Suresh},
  booktitle={IROS},
  year={2024},
}

@article{MWGLBPS23,
  title={From one field to another—Unsupervised domain adaptation for semantic segmentation in agricultural robotics},
  author={Magistri, Federico and Weyler, Jan and Gogoll, Dario and Lottes, Philipp and Behley, Jens and Petrinic, Nik and Stachniss, Cyrill},
  journal={Comput. Electron. Agric.},
  year={2023},
}

@inproceedings{SSN19,
  title={Unsupervised domain adaptation for DNN-based automated harvesting},
  author={Shkanaev, Aleksandr Yu and Sholomov, Dmitry L and Nikolaev, Dmitry P},
  booktitle={ICMV},
  year={2020},
}

@inproceedings{TNG25,
  title={From Web Data to Real Fields: Low-Cost Unsupervised Domain Adaptation for Agricultural Robots},
  author={Tzouras, Vasileios and Nalpantidis, Lazaros and G{\"u}ldenring, Ronja},
  booktitle={SCIA},
  year={2025},
}

@article{GL24,
  title={Unsupervised domain adaptation for highlight detection and removal in agricultural robot vision system.},
  author={Guan, Laide and Li, Bole},
  journal={J. Biotech Res.},
  year={2024}
}

@article{DJPMG22,
  title={Unsupervised domain adaptation using transformers for sugarcane rows and gaps detection},
  author={dos Santos Ferreira, Alessandro and Junior, Jos{\'e} Marcato and Pistori, Hemerson and Melgani, Farid and Gon{\c{c}}alves, Wesley Nunes},
  journal={Comput. Electron. Agric.},
  year={2022},
}

@article{LYHYZA24,
  title={Design and experiment of nighttime greenhouse tomato harvesting robot},
  author={Liu, Lei and Yang, Qizhi and He, Wenbing and Yang, Xinyu and Zhou, Qin and Addy, Min Min and others},
  journal={J. Eng. Technol. Sci.},
  year={2024}
}

@article{WBB24,
  title={Nighttime Harvesting of OrBot (Orchard RoBot)},
  author={Waltman, Jakob and Buchanan, Ethan and Bulanon, Duke M},
  journal={AgriEngineering},
  year={2024},
  publisher={MDPI}
}

@article{TOT25,
  title={Sweet pepper detection in day and night greenhouse environments using thermal and depth imaging: Z. Tasneem et al.},
  author={Tasneem, Zinat and Oka, Koichi and Tada, Naoya},
  journal={ROBOMECH Journal},
  year={2025},
}

@inproceedings{MOYATS17,
  title={Development of long-term night-vision video analyzing system for physical pest control},
  author={Madokoro, Hirokazu and Ohira, Haruki and Yaji, Yukio and Abe, Makoto and Terata, Yuki and Sato, Kazuhito},
  booktitle={SII},
  year={2017},
}

@article{GPMXWOCB20,
  title={Generative adversarial networks},
  author={Goodfellow, Ian and Pouget-Abadie, Jean and Mirza, Mehdi and Xu, Bing and Warde-Farley, David and Ozair, Sherjil and Courville, Aaron and Bengio, Yoshua},
  journal={Communications of the ACM},
  year={2020},
}

@inproceedings{WERHK19,
  title={A {RUGD} dataset for autonomous navigation and visual perception in unstructured outdoor environments},
  author={Wigness, Maggie and Eum, Sungmin and Rogers, John G and Han, David and Kwon, Heesung},
  booktitle={IROS},
  year={2019},
}

@article{PPNZY24,
  title={One-step image translation with text-to-image models},
  author={Parmar, Gaurav and Park, Taesung and Narasimhan, Srinivasa and Zhu, Jun-Yan},
  journal={arXiv},
  year={2024}
}

@article{GKCSWM22,
  title={{GA-Nav}: Efficient terrain segmentation for robot navigation in unstructured outdoor environments},
  author={Guan, Tianrui and Kothandaraman, Divya and Chandra, Rohan and Sathyamoorthy, Adarsh Jagan and Weerakoon, Kasun and Manocha, Dinesh},
  journal={IEEE RA-L},
  year={2022},
}

@inproceedings{SLBR24,
  title={Adversarial diffusion distillation},
  author={Sauer, Axel and Lorenz, Dominik and Blattmann, Andreas and Rombach, Robin},
  booktitle={ECCV},
  year={2024},
}

@inproceedings{LLTLGJXY24,
  title={Light the night: A multi-condition diffusion framework for unpaired low-light enhancement in autonomous driving},
  author={Li, Jinlong and Li, Baolu and Tu, Zhengzhong and Liu, Xinyu and Guo, Qing and Juefei-Xu, Felix and Xu, Runsheng and Yu, Hongkai},
  booktitle={IEEE/CVF CVPR},
  year={2024}
}

@inproceedings{DPNRRPZHWO25,
  title={{M2P2}: A multi-modal passive perception dataset for off-road mobility in extreme low-light conditions},
  author={Datar, Aniket and Pokhrel, Anuj and Nazeri, Mohammad and Rao, Madhan B and Rangwala, Harsh and Pan, Chenhui and Zhang, Yufan and Harrison, Andr{\'e} and Wigness, Maggie and others},
  booktitle={IROS},
  year={2025},
}

@article{LBYSGK15,
  title={Kiwifruit recognition at nighttime using artificial lighting based on machine vision},
  author={Longsheng, Fu and Bin, Wang and Yongjie, Cui and Shuai, Su and Gejima, Yoshinori and Kobayashi, Taiichi},
  journal={Int. J. Agric. Biol. Eng.},
  year={2015}
}

@article{DPBGCCCDDF18,
  title={Agricultural robotics: the future of robotic agriculture},
  author={Duckett, Tom and Pearson, Simon and Blackmore, Simon and others},
  journal={arXiv},
  year={2018}
}

@article{WCCCLSL22,
  title={{SFNet-N}: An improved SFNet algorithm for semantic segmentation of low-light autonomous driving road scenes},
  author={Wang, Hai and Chen, Yanyan and Cai, Yingfeng and Chen, Long and Li, Yicheng and Sotelo, Miguel Angel and Li, Zhixiong},
  journal={IEEE Trans. Intell. Transp. Syst.},
  year={2022},
}

@inproceedings{KB25,
  title={Pixel-aligned rgb-nir stereo imaging and dataset for robot vision},
  author={Kim, Jinnyeong and Baek, Seung-Hwan},
  booktitle={CVPR},
  year={2025}
}

@article{CSNSN24,
  title={Terrain classification method using an NIR or RGB camera with a {CNN}-based fusion of vision and a reduced-order proprioception model},
  author={Chen, Hsiao-Yu and Sang, I-Chen and Norris, William R and Soylemezoglu, Ahmet and Nottage, Dustin},
  journal={Comput. Electron. Agric.},
  year={2024},
}

@article{TAL16,
  title={Resnet in resnet: Generalizing residual architectures},
  author={Targ, Sasha and Almeida, Diogo and Lyman, Kevin},
  journal={arXiv},
  year={2016}
}

@inproceedings{IZZE17,
  title={Image-to-image translation with conditional adversarial networks},
  author={Isola, Phillip and Zhu, Jun-Yan and Zhou, Tinghui and Efros, Alexei A},
  booktitle={CVPR},
  year={2017}
}

@inproceedings{MLXLWP17,
  title={Least squares generative adversarial networks},
  author={Mao, Xudong and Li, Qing and Xie, Haoran and Lau, Raymond YK and Wang, Zhen and Paul Smolley, Stephen},
  booktitle={ICCV},
  year={2017}
}

@article{W22,
  title={Mask {CycleGAN}: Unpaired Multi-modal Domain Translation with Interpretable Latent Variable},
  author={Wang, Minfa},
  journal={arXiv},
  year={2022}
}

@article{DCG23,
  title={A vision-based navigation system for arable fields},
  author={de Silva, Rajitha and Cielniak, Grzegorz and Gao, Junfeng},
  journal={arXiv},
  year={2023}
}

@article{GF15,
  title={Sensing the light environment in plants: photoreceptors and early signaling steps},
  author={Galv{\~a}o, Vinicius Costa and Fankhauser, Christian},
  journal={Curr. Opin. Neurobiol.},
  year={2015},
}

@article{T79,
  title={Red and photographic infrared linear combinations for monitoring vegetation},
  author={Tucker, Compton J},
  journal={Remote Sens. Environ.},
  year={1979},
}

@article{KWBHV21,
  title={Selective harvesting robotics: current research, trends, and future directions},
  author={Kootstra, Gert and Wang, Xin and Blok, Pieter M and Hemming, Jochen and Van Henten, Eldert},
  journal={Curr. Robot. Rep.},
  year={2021},
}
}

\end{document}